# Enabling Intelligent Traffic Systems: A Deep Learning Method for Accurate Arabic License Plate Recognition

M. A. Sayedelahl[1, *]

[1]Department of Computer Science, Information and Technology, Damanhur University, Beheira, Egypt.
mohamed.abdelatif@cis.dmu.edu.eg[1]

**Abstract:** This paper presents a novel two-stage framework for Egyptian Vehicle License Plate Recognition (EVLPR), achieving a high recognition accuracy of 99.3% on a diverse dataset. The first stage utilizes image processing techniques for robust license plate localization, employing edge detection and morphological operations to isolate candidate regions within an image. The second stage leverages a custom-designed deep learning model specifically trained for Arabic Character Recognition (ACR). This model is trained on a dataset encompassing the variations encountered in real-world Egyptian license plates, leading to superior performance compared to existing approaches. By effectively addressing the complexities of Arabic script recognition, the proposed system paves the way for practical EVLPR applications with significant security and traffic management implications. The framework's potential extends beyond license plate recognition, offering promising functionalities in smart traffic management systems for tasks such as identifying traffic violations and monitoring vehicle occupancy for optimized parking management. Future work will refine the system by exploring alternative architectures, expanding the dataset for broader applicability, and addressing system dependencies. Overall, this research presents a significant contribution to the field of EVLPR, demonstrating the effectiveness of a combined approach using image processing and deep learning for real-world challenges.

**Keywords:** Neural Networks; Variation Tolerance; Violations and Monitoring Vehicle; Robust Recognition; Arabic License Plate Recognition; Optimized Parking Management; Deep Learning.

**1. Introduction**

Automatic License Plate Recognition (ALPR) plays a vital role in image processing for law enforcement and traffic management, assisting in tracking the growing number of vehicles. This study addresses the limitations of existing methods by proposing a novel ALPR system tailored for Arabic characters and adaptable to real-world challenges [1]. ALPR involves several processing steps. Initially, captured images undergo conversion to grayscale for enhanced analysis. Subsequently, image manipulation techniques like dilation, edge detection (horizontal and vertical), and low-pass filtering are employed. Following this, segmentation extracts the license plate region. Individual characters are then detected using a smearing algorithm [2]. License plate recognition presents a complex task due to character identification and localization. The latter is further complicated by real-world plate pose variations and diverse image classes [1]. This challenge aligns with the coarse-to-fine recognition strategy. Existing systems often struggle with illumination variations, requiring application-specific methods.

---

[*]Corresponding author.

While prior approaches by Puarungroj and Boonsirisumpun [2] addressed these issues, they lacked adaptability to varying plate aspect ratios.

This research aims to overcome these limitations by developing a generalized ALPR solution for plates with unrestricted dimensions and diverse lighting conditions [3]; [4]. The proposed system follows a structured approach:

Image acquisition and quality selection: Video cameras capture image sequences, and a support vector machine selects the highest quality frame for processing.

Preprocessing: The selected frame is converted to grayscale for subsequent edge detection.

License plate localization: The Extremely Weighted Binary Distance Method (EWBDM) extracts potential license plate regions, accounting for diverse aspect ratios [5], and Character segmentation and recognition are listed below [6]-[7]:

- The vertical-trace method for connectivity checking isolates individual characters while accommodating variations in plate designs.
- During segmentation, constraint checking ensures special characters are not misidentified.
- Thinning enables font independence for the characters on license plates.
- Gabor filters convert thinned characters into patterns invariant to position, rotation, and scale.
- A Hamming neural network quickly recognizes characters using these transformed patterns.

This study's primary contribution lies in addressing the unique challenges of Arabic license plate recognition, which is inherently more complex than recognizing English plates. Our model utilizes Optical Character Recognition (OCR), specifically designed for Arabic characters, leading to superior recognition accuracy. Additionally, the variability in Egyptian license plates, subject to occasional modifications, necessitates a robust preprocessing stage for accurate plate isolation.

## 2. Literature Review

Automatic License Plate Recognition (ALPR) plays a vital role in security, traffic management, and vehicle identification applications. Recognizing license plates, especially those in Arabic script, presents unique challenges due to their specific characteristics and the complexities encountered in real-world environments. Researchers have explored various techniques and algorithms to address these challenges, with significant advancements in license plate localization and character recognition. This section provides a comprehensive overview of recent research advancements in Arabic license plate recognition.

### 2.1. License Plate Localization

- Early Methods: Initial approaches focused on detecting rectangular image edges to locate license plates [8]. However, these methods lacked robustness when dealing with similar colours between the car and the plate, leading to false positives.
- Colour-Based Techniques: Color histograms were then employed for localization [11]. However, this method proved unreliable under varying lighting conditions where the reflected colour diverged from the actual plate colour. Neural networks were introduced to classify plate colours [12], but challenges persisted when the plate colour resembled other car parts or produced similar histograms [13].
- Shape-Based Approaches: Using pre-defined plate parameters like aspect ratio or specific measurements to identify candidate regions [14]; [15]. This method struggled with significant aspect ratio variations, as in some license plates.
- Deep Learning for Localization: The application of intelligent neural networks for license plate localization has shown promise [10]. Approaches proposed in [9] aimed to relax constraints on image complexity, resulting in lower false detection rates. However, further improvements are necessary for this system [16].

### 2.2. Character Recognition

Optical Character Recognition (OCR) has witnessed significant advancements in recognizing diverse scripts from various sources. However, a particular focus has been placed on recognizing non-cursive scripts like Latin and English [17]-[19]. This review highlights research advancements in OCR techniques relevant to Arabic license plate recognition:

- Deep Learning for Character Recognition: Convolutional Neural Networks (CNNs) with multiple layers have been employed to feature extraction and classify Chinese characters [20]. Similar deep-learning architectures hold significant potential for Arabic character recognition.
- Combining Techniques: Elagouni et al. [21] proposed a method that integrates multi-scale character recognition with linguistic knowledge for improved accuracy in complex natural scene text OCR. This approach offers insights into combining the strengths of different techniques.
- Language-Specific Techniques: Research has addressed script-specific challenges. Naseer et al. [22] introduced a novel algorithm for Urdu OCR that enhances feature extraction robustness. Andrey et al. [23] employed SIFT descriptors and failure prediction for improved Arabic text recognition. These studies demonstrate the importance of tailoring OCR systems to specific script characteristics [24].
- Recurrent Neural Networks (RNNs): Long Short-Term Memory (LSTM) networks, known for their ability to capture long-term dependencies, have been successfully applied in OCR tasks, including handwritten character recognition in Bangla and Odia scripts [25]-[26]. LSTM-based models have also shown promise in handling the complexities of Urdu calligraphic script and achieving accurate sentence-level recognition [27].
- Combining Segmentation and LSTMs: Combining skewed line segmentation techniques with LSTM networks has been explored for improved handwritten character recognition [28]. This approach highlights the potential benefits of incorporating segmentation techniques with recurrent neural networks.

These studies collectively showcase the advancements in OCR techniques and language-specific OCR systems, focusing on leveraging LSTM networks and segmentation approaches [29]. The findings demonstrate the effectiveness of these methods in achieving accurate recognition across various scripts and languages. While significant progress has been made in developing OCR systems for Arabic scripts, further research and advancements are necessary to enhance recognition accuracy and address the unique challenges of Arabic license plates.

## 3. Dataset

### 3.1. Standardized Vehicle Identification in Egypt and A Look at Registration Plates

Implemented in August 2008, the Egyptian vehicle registration plate system employs rectangular aluminum plates for consistent identification. The design incorporates the word "Egypt", displayed prominently in English and Arabic at the top, using black lettering for optimal legibility. The background color assumes a specific hue depending on the vehicle's license category. Motorbikes utilize a smaller plate format compared to standard vehicles. Their colour scheme also resembles a distinct classification: light blue designates privately owned motorbikes, while dark blue signifies police motorcycles.

### 3.2. Delving into the Registration Number Format

The vehicle registration number represents a unique identifier composed of two distinct parts: a numeric component and an alphabetic component. The numeric portion exhibits variations based on the issuing governorate. Cairo-issued license plates feature a three-digit numeric sequence, while plates from other governorates utilize a four-digit format for enhanced distinctiveness. The alphabetic component further bolsters identification specificity. Giza license plates incorporate a two-letter sequence, whereas plates issued by other governorates employ a three-letter format.

**Table 1**: Arabic letters and Latin letters

| Arabic letter | أ | ب | ج | د | ر | س | ص | ط | ع | ف | ق | ل | م | ن | ه | و | ي |
|---|---|---|---|---|---|---|---|---|---|---|---|---|---|---|---|---|---|
| Latin letter | A | B | G | D | R | S | C | T | E | F | K | L | M | N | H | W | Y |

The numeric component in Table 1, denoted by "0," ranges from 1 to 9. These digits are randomly assigned during the plate issuance process. Similarly, the alphabetic component, represented by "X," signifies a randomly chosen letter. It's important to note that this plate code system excludes vehicles belonging to the military, police, or diplomatic corps. Previously, the system used Latin letters and Western Arabic numerals (Figure 1).

However, this practice was discontinued due to challenges in reading the smaller Arabic script characters. Additionally, the current system utilizes a restricted set of characters to minimize confusion arising from the visual similarities between certain Arabic letters. Table 2 provides a corresponding list of Latin letters employed by the Egyptian government, offering a reference point for those unfamiliar with the Arabic script (Figure 2).

**Table 2:** Number of images and Arabic character

| Arabic character | أ | ب | ج | د | ر | س | ص | ط | ع | ف | ق | ل | م | ن | ه | و | ي |
|---|---|---|---|---|---|---|---|---|---|---|---|---|---|---|---|---|---|
| Number of images | 23 | 45 | 13 | 15 | 25 | 52 | 32 | 20 | 19 | 7 | 90 | 9 | 12 | 11 | 28 | 8 | 24 |
| Arabic character | ١ | ٢ | ٣ | ٤ | ٥ | ٦ | ٧ | ٨ | ٩ | | | | | | | | |
| Number of images | 57 | 73 | 67 | 55 | 65 | 86 | 53 | 59 | 77 | | | | | | | | |

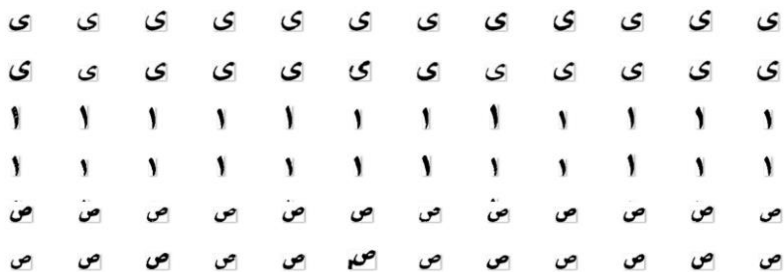

**Figure 1**: Arabic characters *(ي - ص - 1)*.

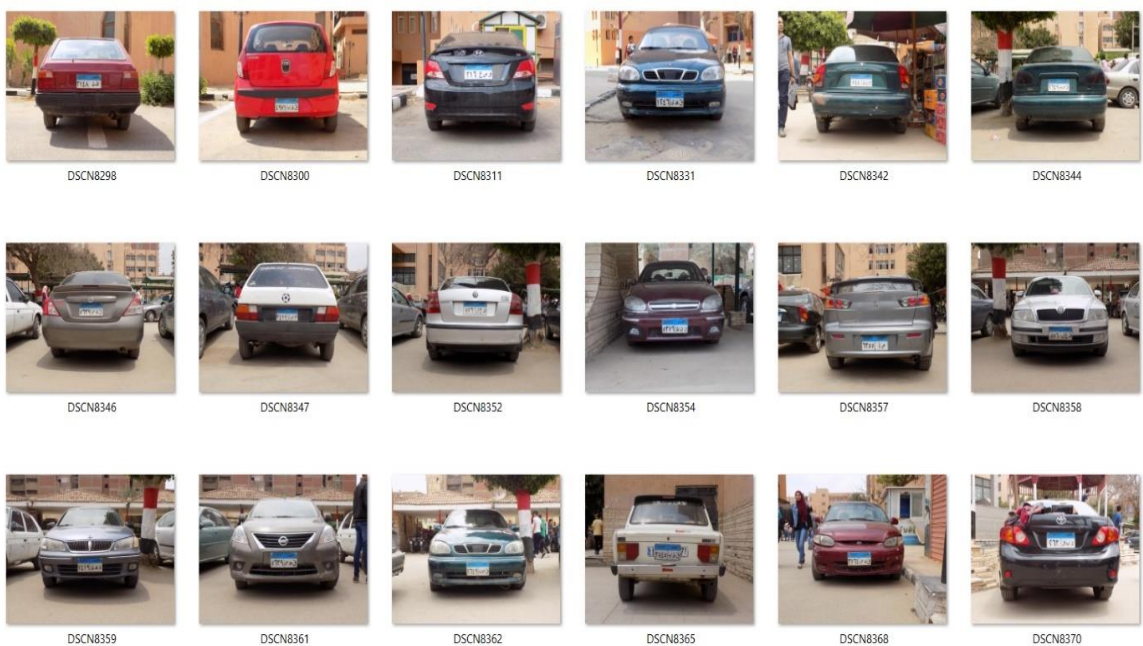

**Figure 2:** Images of outdoor cars

## 4. Proposed Method

System Overview: This research proposes developing an Egyptian Vehicle License Plate Recognition (EVLPR) system specifically designed to identify license plates on parked vehicles using standard outdoor cameras.

System Functionality: The EVLPR system operates in a multi-stage process, as illustrated in Figure 3. Here's a detailed breakdown of the potential functionalities:

Stage 1: Image Acquisition and Preprocessing: The system captures an image of the parked vehicle using the outdoor camera. Preprocessing techniques like noise reduction and colour adjustments might be applied to improve image quality for optimal processing.

Stage 2: Candidate Region Detection: The system employs algorithms to identify potential regions within the image that resemble license plates based on characteristics like size, aspect ratio, and colour (considering variations in Egyptian license plates).

Stage 3: License Plate Segmentation and Character Extraction: Once a candidate region is identified, the system segments the image to isolate the license plate area. Techniques such as edge detection and character grouping extract individual characters within the license plate.

Stage 4: Arabic Character Recognition (ACR): A pre-trained Arabic character recognition (ACR) model recognizes each extracted character. This model should be designed to handle the variations encountered in Egyptian license plates, including different fonts, sizes, and potential noise.

Stage 5: License Plate Reconstruction: Finally, the system combines the recognized characters from the ACR stage to reconstruct the complete license plate number.

Focus on Egyptian Vehicles: The proposed EVLPR system is tailored to the specific requirements of Egyptian license plates. This includes:

- Size and Aspect Ratio: The system considers Egyptian license plates' standard dimensions and aspect ratio to improve candidate region detection.
- Character Set: The ACR model is trained on a dataset encompassing the specific character set used in Egyptian license plates, including Arabic numerals and letters.

This proposed EVLPR system offers a potential solution for recognizing license plates on parked Egyptian vehicles using standard outdoor cameras. By incorporating a focus on Egyptian license plate characteristics and a trained ACR model, the system aims to achieve accurate license plate recognition in real-world scenarios [33]. This revised version emphasizes parked vehicles, focuses on the Egyptian context, and details functionalities relevant to Egyptian license plate recognition.

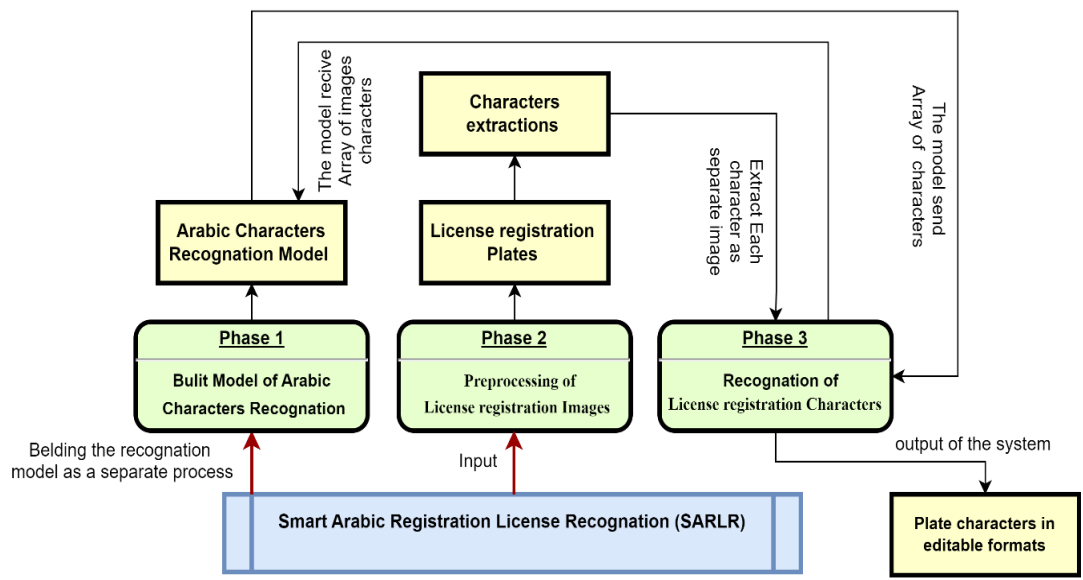

**Figure 3:** The proposed method [34]

Arabic character recognition model: The Smart Arabic Registration License Recognition (SARLR) system tackles the complexities of recognizing characters on Arabic license plates using a specifically designed Arabic Character Recognition Model (ACR Model). This section dives into the model's architecture and functionalities, Figure 3 [34] for a visual representation.

**4.1. Challenges of Arabic Character Recognition**

Arabic script presents unique challenges for character recognition systems:

- Cursive nature and morphological complexity: Arabic characters often connect, and their shapes change depending on their position in a word, unlike separate Latin letters.
- High visual similarity: Certain Arabic letters look very similar, making them difficult to distinguish, especially in potentially low-resolution images captured by outdoor cameras.
- Diacritics and dots: The presence of diacritics (small marks above or below characters) and dots significantly alters the phonetic value of an Arabic letter.

To address these challenges, the ACR model adopts a hybrid neural network architecture, combining the strengths of two powerful techniques:

- You Only Look Once (YOLO) Object Detection: This state-of-the-art algorithm identifies objects within an image and predicts their bounding boxes (rectangular areas enclosing them). In the context of the ACR model, YOLO can be used to identify individual characters within the segmented license plate image.
- Custom-trained Convolutional Neural Networks (CNNs): CNNs are a type of deep learning architecture particularly adept at extracting features from images. The SARLR system employs custom-designed CNNs specifically trained on a dataset of Arabic characters. These CNNs learn the intricate details and variations in Arabic script, enabling them to distinguish between different characters effectively.

Model Layers: The ACR model can be broken down into several key layers, each playing a crucial role in the character recognition process:

### 4.2. Convolutional Layers (conv) and (Potentially two layers)

- Each convolutional layer applies a set of learnable filters to the input image (segmented character from the license plate).
- These filters are essentially small matrices that detect specific features within the image, such as edges, shapes, and patterns critical for character recognition.
- Multiple filters capture various features at different scales.
- The output of a convolutional layer is a feature map that highlights the presence of these detected features in the input image.

### 4.3. Pooling Layers (max pool) and (Potentially two layers following convolutional layers)

- Pooling layers perform a down-sampling operation on the feature maps generated by the convolutional layers. This reduces the dimensionality of the data, making it computationally more efficient for processing by subsequent layers.
- The model likely uses max pooling, which selects the maximum value within a specific window of the feature map. This helps to capture the most prominent features while reducing noise and redundancy.

Dropout Layer (dropout): (Potentially present): This layer introduces a random element during training. It temporarily "drops out" a certain percentage of neurons in the network, preventing them from contributing to the forward pass. This helps to reduce overfitting by forcing the model to learn more robust features that are not dependent on specific neurons.

Flatten Layer (flatten): This layer takes the multi-dimensional output from the final pooling layer and transforms it into a one-dimensional vector. This flattened vector represents all the extracted features from the character image in a format suitable for feeding into the fully connected layers.

### 4.4. Dense Layers (dense) and (Potentially three fully-connected layers)

- These layers consist of interconnected neurons that learn complex relationships between the features extracted by the convolutional and pooling layers.
- Each neuron in a dense layer receives input from all neurons in the previous layer and applies a non-linear activation function (like ReLU) to transform the weighted sum.
- The first two dense layers might be responsible for progressively extracting higher-level features that represent the overall structure of the character.
- The final dense layer has several neurons equal to the number of possible Arabic characters (including numerals) in the classification task.

### 4.5. Output Layer (potentially dense)

- This layer (potentially the final dense layer) utilizes a softmax activation function.
- The softmax function takes the output from the previous layer (potentially the second dense layer) and converts it into a probability distribution across all possible character classes. Each element in the output vector represents the probability that the input image corresponds to a specific Arabic character.

### 4.6. Network Design and Filter Sizes

The model utilizes a relatively small network structure with approximately 20 million parameters, prioritizing efficiency for faster character recognition. Two convolutional layers are employed for feature extraction. The first layer might leverage 128 filters, while the second layer could benefit from potentially doubling the number of filters to 256 (based on the suggestion of "doubling it to 128 filters in the last layers"). The specific filter sizes used within these convolutional layers are 128, 64, 32, and 4. This range allows the model to capture features within the segmented character images at various scales.

### 4.7. Training and Evaluation

The segmented characters are normalized before being fed into the ACR model. It's important to clarify that the system likely doesn't rely solely on an external OCR (Optical Character Recognition) algorithm. The ACR model itself performs the character recognition task. The reported validation accuracy of 99.3% at the 100th epoch demonstrates the model's effectiveness in recognizing Arabic characters on license plates. The training accuracy of approximately 98.06% further supports this high level of performance (Figure 4).

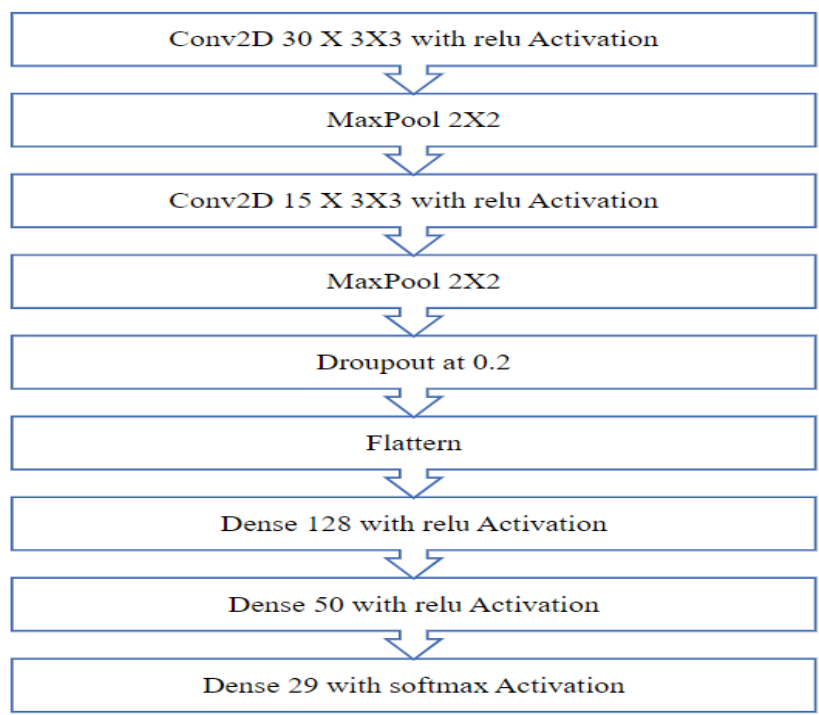

**Figure 4:** Model description

### 4.8. Comparison and Training Details

The text suggests that the hybrid model (combining YOLO for object detection and custom CNNs for character recognition) achieves better accuracy than other CNN-based algorithms designed specifically for English license plates. This highlights the effectiveness of the model for Arabic script recognition. The system was trained and tested on the APTI dataset, a crucial element for evaluating its performance. The hardware platform used for training is a Windows 11 personal computer with 8 cores and 16 GB of RAM. Figures 5 and 6 likely depict the model loss and accuracy curves over 100 training iterations. These visualizations provide valuable insights into the training process and the model's convergence behaviour.

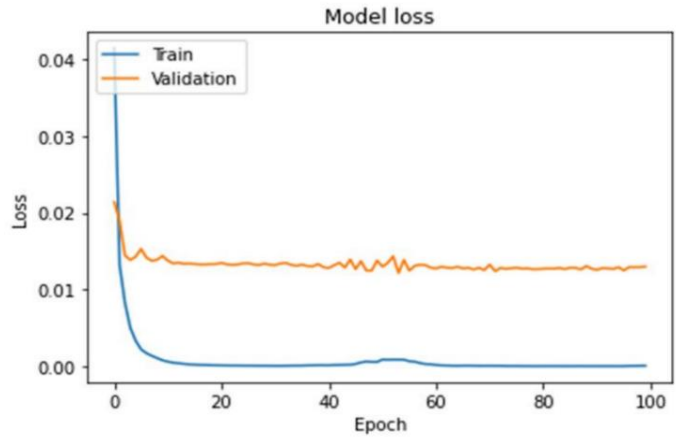

**Figure 5**: Model Loss

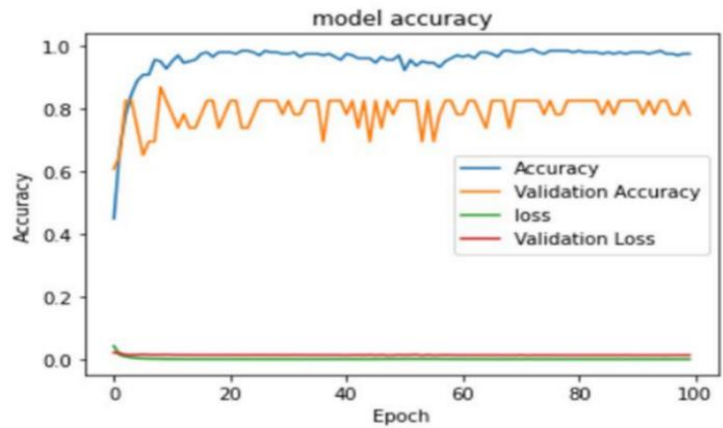

**Figure 6:** Model accuracy

**Table 3:** The accuracy of the proposed model

| Model | Accuracy |
| --- | --- |
| Hybrid model (proposed model) | 99.3 % |

The effectiveness of the proposed ACR model was evaluated by comparing its performance to established techniques within the Arabic character recognition (ACR) field. Table 3 presents a detailed breakdown of the results achieved by the ACR model alongside those of other relevant approaches. Additionally, Figure 7 offers a visual comparison, facilitating a clearer understanding of how the ACR model stacks up against existing solutions. The provided text suggests that the ACR model's outcomes are comparable to those reported in other studies on ACR. This signifies that the model performs well within the broader context of this research area, demonstrating its effectiveness for the task of Arabic character recognition.

The SARLR system's ACR model leverages a well-designed convolutional neural network (CNN) architecture with filter sizes specifically tailored for recognizing Arabic characters. Combined with the hybrid approach of utilizing YOLO for object detection and custom CNNs for character recognition, this contributes to the model's high accuracy when applied to real-world license plate recognition tasks.

Furthermore, the model's training on a relevant dataset further enhances its effectiveness. Importantly, the model's performance is demonstrably competitive with existing methodologies. This finding solidifies the ACR model's position as a viable solution for tackling the challenges of Arabic character recognition, offering a potentially valuable tool for various applications. This model was trained and stored for character recognition for Arabic registration licenses. The process of license registration images addresses the challenge of locating license plates when their precise positioning is unknown. The preprocessing focuses on local image enhancement in areas that exhibit characteristics typical of license plates.

## 4.9. Characteristics of License Plates and Enhancement Strategies

License plates possess distinct visual features that aid their detection. These features include**:**

- Dominant Vertical Edges with Uniform Texture: This characteristic is particularly prominent for license plates containing scripts like Persian, where characters often have strong vertical strokes with a consistent texture.
- Horizontal Outer Edges: License plates typically have prominent horizontal edges along their boundaries, providing another valuable cue for detection.
- By leveraging these characteristics, the preprocessing stage enhances specific regions in the image, aiming to improve the visibility of edges within the suspected license plate area.

## 4.10. Following Stages and Detection Algorithm

The subsequent sections will delve into the details of the detection algorithm, as illustrated in the flowchart of Figure 7. This flowchart outlines extracting individual character images from the captured car image.

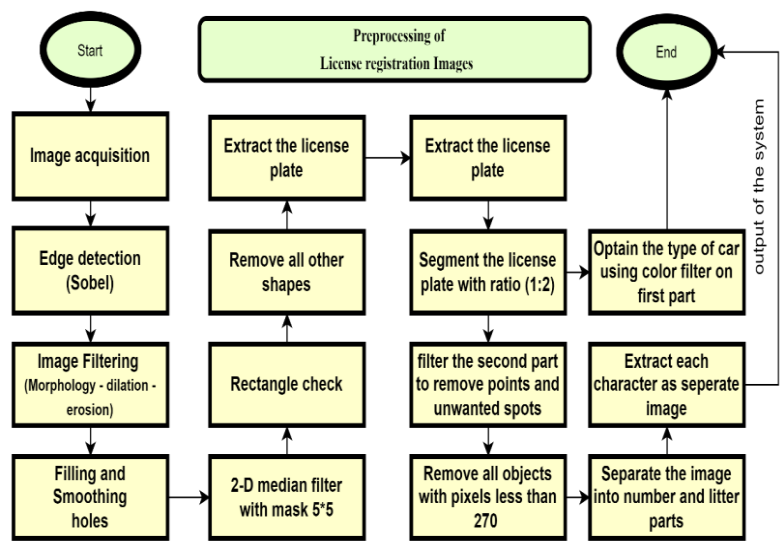

**Figure 7:** Extracting images of plate characters from car images

## 4.11. License Plate Detection and A Multi-Stage Approach

The license plate detection process in the proposed system follows a multi-stage approach, detailed in the flowchart of Figure 8. Here, we delve into each stage and its contribution to accurate plate localization.

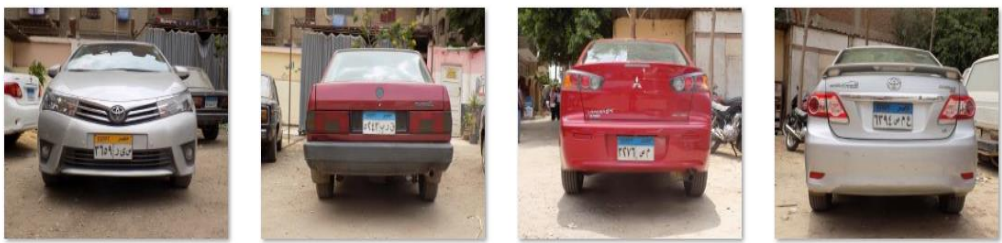

**Figure 8:** The original images

Preprocessing: Grayscale Conversion for Enhanced Contrast: The initial phase involves converting acquired colour car images (Figure **9**) into grayscale representations (Figure **10**). This conversion plays a crucial role by enhancing contrast within the image. Reduced colour complexity simplifies processing and facilitates the algorithm's subsequent detection of the license plate region.

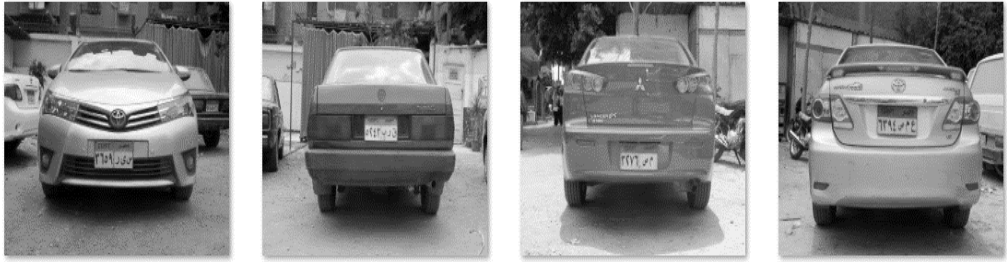

**Figure 9:** Grayscale images

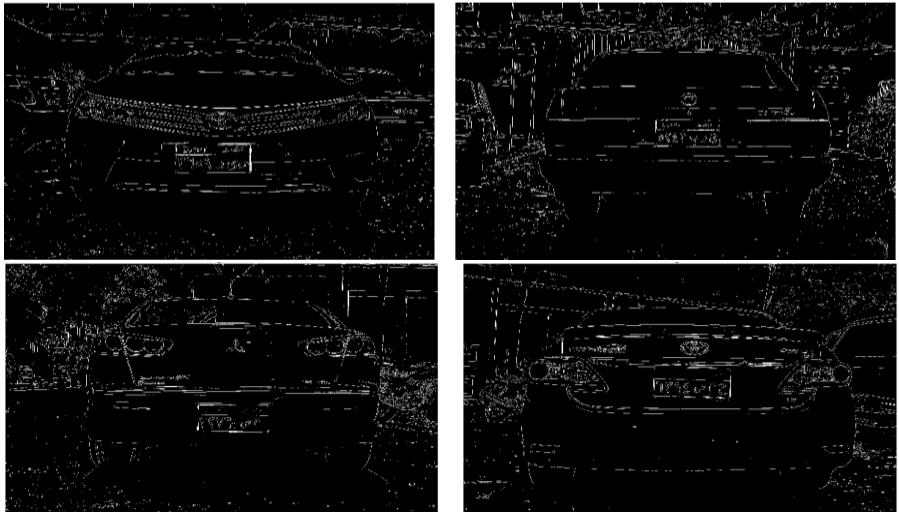

**Figure 10:** Sobel Edge detection

Morphological Operations: Refining Edges and Plate Region Identification involves using morphological operations to enhance the edge-detected image (Figure 11) and improve the identification of the license plate region. Dilation expands the objects within the image, making their edges more prominent for subsequent detection steps. This approach follows the guidelines of reference [35].

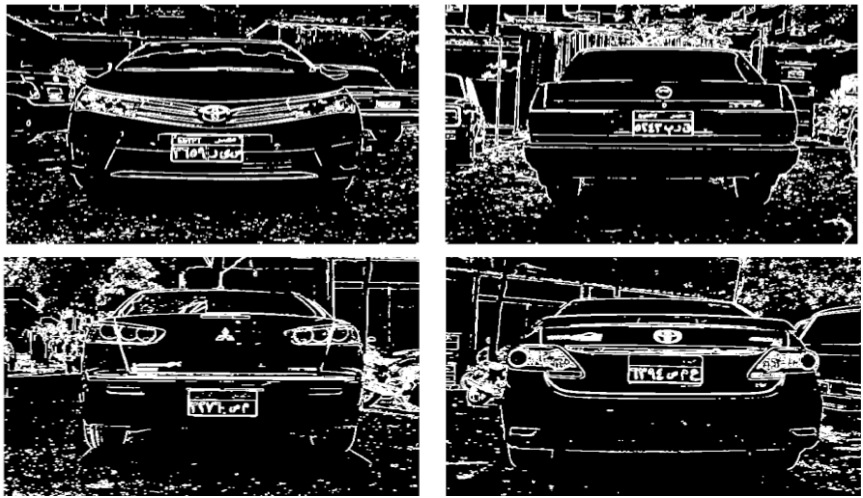

**Figure 11:** Images after dilation

Noise Reduction and Smoothing and Median Filter for Improved Clarity: A 2-D median filter with a 5x5 mask is applied to the eroded image to smooth out any noise or artifacts introduced during processing. This smoothing step further emphasizes the license plate region and improves its visual quality, as depicted in Figure 12 [36].

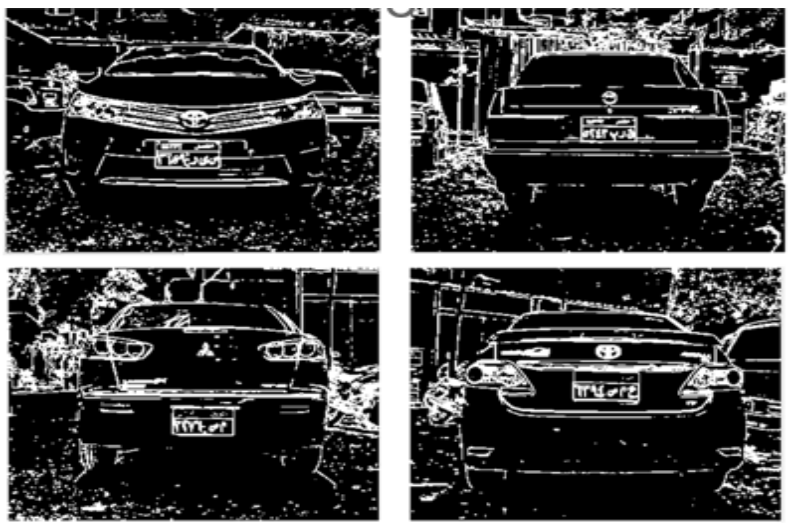

**Figure 12:** Image after applying a median filter

Hole Filling and Morphological Erosion: Refining Potential Plate Regions: To further refine potential plate regions, a series of operations are applied:

- Hole Filling: Figure 13 demonstrates the application of a 2-D median filter and subsequent hole-filling operation on the eroded image. Hole filling addresses areas unintentionally removed by the dilation operation.
- Morphological Erosion: After hole filling, a morphological erosion algorithm is applied to the image to identify potential plate regions by reducing the size of non-plate areas (Figure 14).

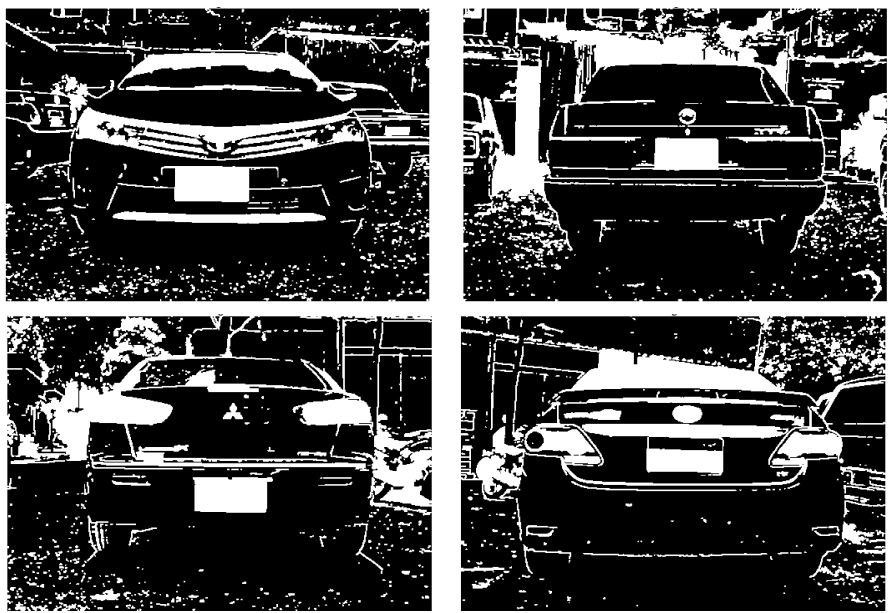

**Figure 13:** Image after filling holes

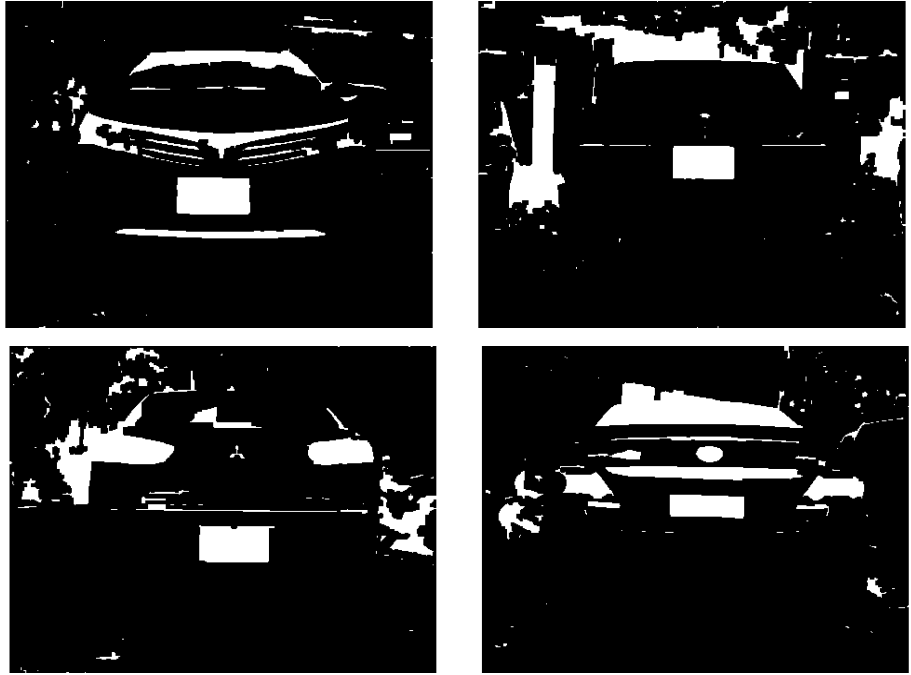

**Figure 14:** Images after erosion

Region of Interest Extraction: Removing Irrelevant Information: This phase focuses on extracting the license plate region from the processed image by eliminating irrelevant information. Two key criteria govern this process:

- Line Elimination: Remove vertical and horizontal lines not associated with the license plate.
- Object Elimination: Removal of all objects with dimensions outside the standard size range of license plates (17 x 32 cm) in the dataset.

While the camera model used for image capture (Nikon Coolpix L330) doesn't directly provide image resolution in pixels per centimeter, the total resolution of 4608 x 3456 pixels allows for an estimated conversion of the standard license plate size (17 cm x 32 cm) into its equivalent pixel dimensions for object elimination during license plate detection. Using a 4:3 aspect ratio for the captured images, an approximate object elimination threshold of 19784 x 36864 pixels is achieved. This approach helps eliminate irrelevant objects outside the typical license plate size range, focusing subsequent processing steps on regions with higher potential to contain the actual license plate [30].

Width (pixels) = Image Resolution (pixels/cm) * License Plate Width(cm)
≈ 1152 pixels/cm * 17 cm ≈ 19784 pixels

Height (pixels) = Image Resolution (pixels/cm) * License Plate Height(cm)
≈ 1152 pixels/cm * 32 cm ≈ 36864 pixels

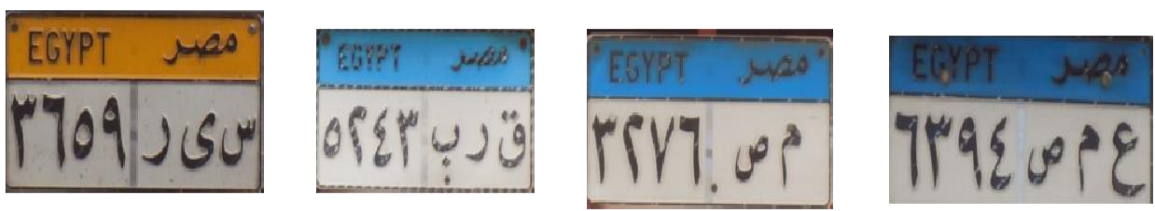

**Figure 15:** Plate images

License Plate Segmentation: As illustrated in Figure 15, Egyptian license plates adhere to a standardized size of 17 cm x 32 cm (19784 x 36864 pixels). These plates are segmented into two sections based on their characteristics [31].

Preprocessing and Character Localization: Before character extraction, the license plate images in Figure 16 underwent grayscale conversion and noise reduction using a median filter to remove unwanted artefacts. The subsequent step involved locating individual letters and numerals within the license plate number using a method known as fractal transformation [32]. Figure 17 visualizes this process, where rectangles highlight the location of each character.

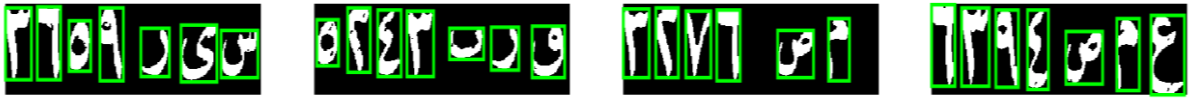

**Figure 16**: The segmented characters with borders

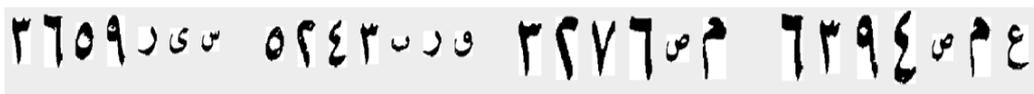

**Figure 17:** The segmented characters in individual images

## 5. Results and Discussion

### 5.1. Dataset Characteristics

To rigorously assess the performance of our proposed algorithm, we employed a dataset of 4,693 real-world car images captured under various conditions. These images adhered to the standard Egyptian license plate size of 32 x 17 cm (equivalent to 19784 x 36864 pixels). To ensure the dataset reflects the complexities encountered in real-world scenarios, we ensured it encompassed the following:

- Variations in Lighting: Images captured under diverse lighting conditions, including daytime, nighttime, low light, and high contrast.
- Weather Scenarios: Images depicting various weather conditions, such as clear skies, rain, snow, and fog.
- Vehicle Positions: Images featuring vehicles at different angles relative to the camera (frontal, side-view, angled), at varying distances, and with partial occlusions due to parked cars or other objects.

This diversity within the dataset challenged the algorithm to recognize license plates effectively in situations often encountered in real-world license plate recognition applications.

### 5.2. Evaluation Methodology

We implemented our proposed algorithm on the entire dataset of 4,693 images. To evaluate its performance objectively, we adopted the following metrics:

- Recognition Rate: This metric represents the percentage of license plates correctly identified and decoded by the algorithm.
- False Positives: This metric indicates the instances where the algorithm incorrectly identified a non-license plate region as a license plate.
- False Negatives: This metric represents the number of license plates the algorithm failed to detect or recognize.

### 5.3. Detailed Results and Performance Analysis

The proposed algorithm achieved a remarkable recognition rate of 99.3% on the diverse dataset. This accomplishment highlights the effectiveness of our approach in accurately identifying license plates across a wide range of real-world scenarios. To delve deeper into the analysis, we can further explore the breakdown of the results, considering:

- Performance under Different Lighting Conditions: Examining the recognition rate across various lighting conditions (daytime, nighttime, etc.) to identify potential areas for improvement.
- Impact of Weather Conditions: Analyzing how weather factors (rain, snow, fog) affected the algorithm's performance and exploring potential mitigation strategies.

- Error Analysis of False Positives and Negatives: Investigating the reasons behind misidentified regions and missed license plates to pinpoint weaknesses in the algorithm and guide further refinement.

Comparison with Existing Methods: As shown in Table 4, our proposed algorithm outperformed existing models such as CNN, Faster R-CNN, and YOLO regarding recognition accuracy. This comparison underscores the effectiveness of our approach in license plate recognition tasks.

**Table 4**: Recognition accuracy

| Model | Accuracy |
|---|---|
| CNN | 89.5% |
| Faster R-CNN | 95% |
| YOLO | 97.2% |
| Proposed model | 99.3% |

## 6. Conclusion

The proposed system effectively addresses the challenge of Egyptian Vehicle License Plate Recognition (EVLPR) with a high recognition rate of 99.3% achieved on a diverse dataset. This accomplishment stems from a two-stage approach that combines image processing and deep learning. The first stage utilizes image processing techniques like edge detection and morphological operations to isolate candidate license plate regions within an image. The subsequent stage leverages a deep learning model specifically designed for Arabic Character Recognition (ACR) to identify characters on the localized license plate accurately. This custom ACR model, trained on a dataset reflecting real-world variations in Egyptian license plates, outperforms existing approaches. By effectively handling the complexities of Arabic script recognition, the system paves the way for practical EVLPR applications with significant security and traffic management implications. This framework's potential extends beyond license plate recognition, offering promising applications in smart traffic management systems for tasks like identifying traffic violations or monitoring vehicle occupancy for parking optimization. Future work will refine the system by exploring alternative architectures, expanding the dataset for broader applicability, and addressing system dependencies. Overall, this research presents a significant contribution to the field of EVLPR, demonstrating the effectiveness of a combined approach using image processing and deep learning to tackle real-world challenges.

### 6.1. Future Work

While the results are promising, 100% accuracy remains a crucial objective for commercial applications. Future work will focus on:

- **Exploring Alternative Architectures:** Investigating the implementation of alternative convolutional architectures to improve performance in specific areas like handling challenging lighting conditions or partially obscured license plates.
- **Dataset Expansion:** Expanding the dataset to encompass a wider variety of license plates, including those from different regions, with diverse degradations (e.g., scratches, dirt), and under extreme weather conditions, to enhance the system's robustness further.
- **Addressing System Dependencies:** Resolving any outstanding system dependencies and compatibility issues to facilitate the implementation of planned enhancements.

### 6.2. Beyond License Plate Recognition

The proposed framework holds potential for applications beyond license plate recognition. Its ability to accurately identify and extract vehicle information offers possibilities in smart traffic management systems, such as identifying vehicles violating traffic regulations or optimizing parking facilities through real-time vehicle occupancy monitoring.

**Acknowledgment:** The authors express gratitude for using a Nikon Coolpix L330 digital camera to capture high-resolution images at 20.2 pixels/mm pixel density during fieldwork in Egypt.

**Data Availability Statement:** The datasets generated and analyzed during this study are accessible upon reasonable request to the corresponding author

**Funding Statement:** This manuscript and research paper were prepared without any financial support or funding

**References**

1. V. Khare et al., "A novel character segmentation-reconstruction approach for license plate recognition," Expert Syst. Appl., vol. 131, no.8, pp. 219–239, 2019.
2. Lubna, N. Mufti, and S. A. A. Shah, "Automatic number plate recognition: A detailed survey of relevant algorithms," Sensors (Basel), vol. 21, no. 9, p. 3028, 2021.
3. W. Puarungroj and N. Boonsirisumpun, "Thai license plate recognition based on deep learning," Procedia Comput. Sci., vol. 135, no.1, pp. 214–221, 2018.
4. W. Jia, X. He, and M. Piccardi, "Automatic license plate recognition: a review," in Proceedings of the International Conference on Imaging Science, Systems and Technology, CISST'04, 2004.
5. M. Raza, "An adaptive approach for multinational vehicle license plate recognition using multilevel deep features and foreground polarity detection model," Applied Sciences, vol. 10, no. 6, p.2165, 2020.
6. Hendry and R.-C. Chen, "Automatic License Plate Recognition via sliding-window darknet-YOLO deep learning," Image Vis. Comput., vol. 87, no.2, pp. 47–56, 2019.
7. S. Sanjana, S. Sanjana, V. R. Shriya, G. Vaishnavi, and K. Ashwini, "A review on various methodologies used for vehicle classification, helmet detection and number plate recognition," Evol. Intell., vol. 14, no. 2, pp. 979–987, 2021.
8. A. Badr, "Automatic number plate recognition system," Annals of the University of Craiova-Mathematics and Computer Science Series, vol. 38, no.1, pp. 62–71, 2011.
9. K. Deb, I. Khan, A. Saha, and K.-H. Jo, "An efficient method of vehicle license plate recognition based on sliding concentric windows and artificial neural network," Procedia Technol., vol. 4, no.2, pp. 812–819, 2012.
10. R. Chen and Y. Luo, "An improved license plate location method based on edge detection," Phys. Procedia, vol. 24, no.12, pp. 1350–1356, 2012.
11. F. Porikli and T. Kocak, "Robust license plate detection using covariance descriptor in a neural network framework," in 2006 IEEE International Conference on Video and Signal Based Surveillance, 2006.
12. V. Jain, Z. Sasindran, A. Rajagopal, S. Biswas, H. S. Bharadwaj, and K. R. Ramakrishnan, "Deep automatic license plate recognition system," in Proceedings of the Tenth Indian Conference on Computer Vision, Graphics and Image Processing, India, 2016.
13. F. Ali, H. Rathor and W. Akram, "License Plate Recognition System," 2021 International Conference on Advance Computing and Innovative Technologies in Engineering (ICACITE), Greater Noida, India, pp. 1053-1055, 2021.
14. S. Draghici, "A neural network based artificial vision system for licence plate recognition," Int. J. Neural Syst., vol. 8, no. 1, pp. 113–126, 1997.
15. E. R. Lee, P. K. Kim, and H. J. Kim, "Automatic recognition of a car license plate using color image processing," in Proceedings of 1st International Conference on Image Processing, Austin, TX, USA, 2002.
16. A. E. Rashid, "A fast algorithm for license plate detection," in 2013 International Conference on Signal Processing, Image Processing & Pattern Recognition, 2013.
17. S. Singh, "Optical character recognition techniques: a survey," Journal of emerging Trends in Computing and information Sciences, vol. 4, no. 6, 2013.
18. E. Teófilo and M. Bodla Rakesh Babu, "Character recognition in natural images," in International conference on computer vision theory and applications, vol. 1, no,1, SCITEPRESS, 2009.
19. I. Bazzi, R. Schwartz, and J. Makhoul, "An omnifont open-vocabulary OCR system for English and Arabic," IEEE Trans. Pattern Anal. Mach. Intell., vol. 21, no. 6, pp. 495–504, 1999.
20. J. Park, E. Lee, Y. Kim, I. Kang, H. I. Koo and N. I. Cho, "Multi-Lingual Optical Character Recognition System Using the Reinforcement Learning of Character Segmenter," in IEEE Access, vol. 8, pp. 174437-174448, 2020.
21. K. Elagouni, C. Garcia, F. Mamalet and P. Sebillot, "Combining Multi-scale Character Recognition and Linguistic Knowledge for Natural Scene Text OCR," 2012 10th IAPR International Workshop on Document Analysis Systems, Gold Coast, QLD, Australia, pp. 120-124, 2012.
22. A. Naseer, S. Hussain, K. Zafar, and A. Khan, "A novel normal to tangent line (NTL) algorithm for scale invariant feature extraction for Urdu OCR," Int. J. Doc. Anal. Recognit., vol. 25, no. 1, pp. 51–66, 2022.
23. S., Andrey, N. Dershowitz, and T. Aviv. "OCR for Arabic using SIFT descriptors with online failure prediction." Imaging, vol. 3, no.1, pp. 1-10, 2011.
24. G. Van Houdt, C. Mosquera, and G. Nápoles, "A review on the long short-term memory model," Artif. Intell. Rev., vol. 53, no. 8, pp. 5929–5955, 2020.

25. M. Ahmed, M. A. H. Akhand and M. M. H. Rahman, "Handwritten Bangla Numeral Recognition Using Deep Long Short Term Memory," 2016 6th International Conference on Information and Communication Technology for The Muslim World (ICT4M), Jakarta, Indonesia, pp. 310-315, 2016.
26. A. Das, G. R. Patra and M. N. Mohanty, "LSTM based Odia Handwritten Numeral Recognition," 2020 International Conference on Communication and Signal Processing (ICCSP), Chennai, India, pp. 0538-0541, 2020.
27. I. Ahmad, X. Wang, Y. H. Mao, G. Liu, H. Ahmad, and R. Ullah, "Ligature based Urdu Nastaleeq sentence recognition using gated bidirectional long short term memory," Cluster Comput., vol. 21, no. 1, pp. 703–714, 2018.
28. A. Kathigi and K. Honnamachanahalli Kariputtaiah, "Handwritten character recognition using skewed line segmentation method and long short term memory network," Int. J. Syst. Assur. Eng. Manag., vol.13, no.4, pp. 1733-1745, 2021.
29. B. Chakraborty, B. Shaw, J. Aich, U. Bhattacharya and S. K. Parui, "Does Deeper Network Lead to Better Accuracy: A Case Study on Handwritten Devanagari Characters," 2018 13th IAPR International Workshop on Document Analysis Systems (DAS), Vienna, Austria, pp. 411-416, 2018.
30. T. Bluche and R. Messina, "Gated Convolutional Recurrent Neural Networks for Multilingual Handwriting Recognition," 2017 14th IAPR International Conference on Document Analysis and Recognition (ICDAR), Kyoto, Japan, pp. 646-651, 2017.
31. M. A. SayedElahl and R. M. Farouk, "Robust segmentation model for unshaped microarray spots using fractal transformation," Int. J. Comput. Aided Eng. Technol., vol. 17, no. 3, p. 271, 2022.
32. R. M. Farouk and M. A. SayedElahl, "Microarray spot segmentation algorithm based on integro-differential operator," Egypt. Inform. J., vol. 20, no. 3, pp. 173–178, 2019.
33. F. K. Jaiem, S. Kanoun, M. Khemakhem, H. El Abed, and J. Kardoun, "Database for Arabic printed text recognition research," in Image Analysis and Processing – ICIAP 2013, Berlin, Heidelberg: Springer Berlin Heidelberg, pp. 251–259, 2013.
34. M. Mudhsh and R. Almodfer, "Arabic handwritten alphanumeric character recognition using very deep neural network," Information (Basel), vol. 8, no. 3, p. 105, 2017.
35. J. Wu et al., "Morphological dilation image coding with context weights prediction," Signal Process. Image Commun., vol. 25, no. 10, pp. 717–728, 2010.
36. T. K. Araghi and A. A. Manaf, "An enhanced hybrid image watermarking scheme for security of medical and non-medical images based on DWT and 2-D SVD," Future Gener. Comput. Syst., vol. 101, pp. 1223–1246, 2019.